\documentclass[10pt,twocolumn,letterpaper,dvipsnames,table]{article}
\usepackage{cvpr}              

\definecolor{cvprblue}{rgb}{0.21,0.49,0.74}
\usepackage[pagebackref,breaklinks,colorlinks,allcolors=cvprblue]{hyperref}

\usepackage[inline]{enumitem}
\usepackage{multirow}
\usepackage{graphicx}
\usepackage{amsmath}
\usepackage{amssymb}
\usepackage{booktabs}
\usepackage{siunitx}
\usepackage{arydshln}
\usepackage[accsupp]{axessibility}

\definecolor{Gray}{gray}{0.9}
\definecolor{input_space}{RGB}{17,119,51}
\definecolor{embedding_space}{RGB}{136,204,238}

\newcommand\blfootnote[1]{%
  \begingroup
  \renewcommand\thefootnote{}\footnote{#1}%
  \addtocounter{footnote}{-1}%
  \endgroup
}

\def\paperID{16837} 
\def\confName{CVPR}
\def\confYear{2025}

\title{\vspace{-20pt}The PanAf-FGBG Dataset:\\ Understanding the Impact of Backgrounds in Wildlife Behaviour Recognition\vspace{-18pt}}

\author{Otto Brookes$^{1,2\dag*}$,\hspace{0.2em}Maksim Kukushkin$^{3,4*}$,\hspace{0.2em}Majid Mirmehdi$^{1}$,\hspace{0.2em}Colleen Stephens$^{5}$,\hspace{0.2em}Paula Dieguez$^{5}$,\vspace{0.1em}\\
\hspace{0.2em}Thurston C. Hicks$^{6}$,\hspace{0.2em}Sorrel Jones$^{5}$,\hspace{0.2em}Kevin Lee$^5$,\hspace{0.2em}Maureen S. McCarthy$^{5}$,\vspace{0.1em}\hspace{0.2em}Amelia Meier$^{5}$,\\
\hspace{0.2em}Emmanuelle Normand$^2$,\hspace{0.2em}Erin G. Wessling$^{7}$\vspace{0.1em},\hspace{0.2em}Roman M.Wittig$^{5,8}$,\hspace{0.2em}Kevin Langergraber$^{9}$,\\
\hspace{0.2em}Klaus Zuberbühler$^{10}$,\hspace{0.2em}Lukas Boesch$^{2}$,\hspace{0.2em}Thomas Schmid$^{3,11}$,\hspace{0.2em}Mimi Arandjelovic$^{5}$,\vspace{0.1em} \\
\hspace{0.2em}Hjalmar Kühl$^{5,12}$,\hspace{0.2em}Tilo Burghardt$^{1}$\vspace{0.2em} \\
{\normalsize $^1$University of Bristol,}
{\normalsize $^2$Wild Chimpanzee Foundation,}
{\normalsize $^{3}$Martin Luther University Halle-Wittenberg,}
\\
{\normalsize $^{4}$Leipzig University,}
{\normalsize $^{5}$Max Planck Institute for Evolutionary Anthropology,} 
{\normalsize $^{6}$University of Warsaw,}
\\
{\normalsize $^{7}$Harvard University,}
{\normalsize $^{8}$University of Lyon,} 
{\normalsize $^{9}$Arizona State University,} 
{\normalsize $^{10}$University of St Andrews,}
\\
{\normalsize $^{11}$Lancaster University in Leipzig,} 
{\normalsize $^{12}$Senckenberg Museum of Natural History }
\\
}

\begin{document}
\maketitle
\begin{abstract}

Computer vision analysis of camera trap video footage is essential for wildlife conservation, as captured behaviours offer some of the earliest indicators of changes in population health. Recently, several high-impact animal behaviour datasets and methods have been introduced to encourage their use; however, the role of behaviour-correlated background information and its significant effect on out-of-distribution generalisation remain unexplored. In response, we present the PanAf-FGBG dataset, featuring 21 hours of wild chimpanzee behaviours, recorded at 389 individual camera locations. Uniquely, it pairs every video with a chimpanzee (referred to as a foreground video) with a corresponding background video (with no chimpanzee) from the same camera location. We present two views of the dataset: one with overlapping camera locations and one with disjoint locations. This setup enables, for the first time, direct evaluation of in-distribution and out-of-distribution conditions, and for the impact of backgrounds on behaviour recognition models to be quantified. All clips have rich behavioural annotations and metadata, including unique camera IDs. Additionally, we establish several baselines and present a highly effective latent-space normalisation technique that boosts out-of-distribution performance by +5.42\% mAP for convolutional and +3.75\% mAP for transformer-based models. Finally, we provide an in-depth analysis on the role of backgrounds in out-of-distribution behaviour recognition, including the so far unexplored impact of background durations. (i.e., the count of background frames within foreground videos). The dataset is available at \href{https://obrookes.github.io/panaf-fgbg.github.io/}{https://obrookes.github.io/panaf-fgbg.github.io/}.
\end{abstract}

\noindent\blfootnote{
    \dag Corresponding author, *Equal technical contribution
}

\begin{figure}[!ht]
\centering
\includegraphics[width=1.0\linewidth, height=4in]{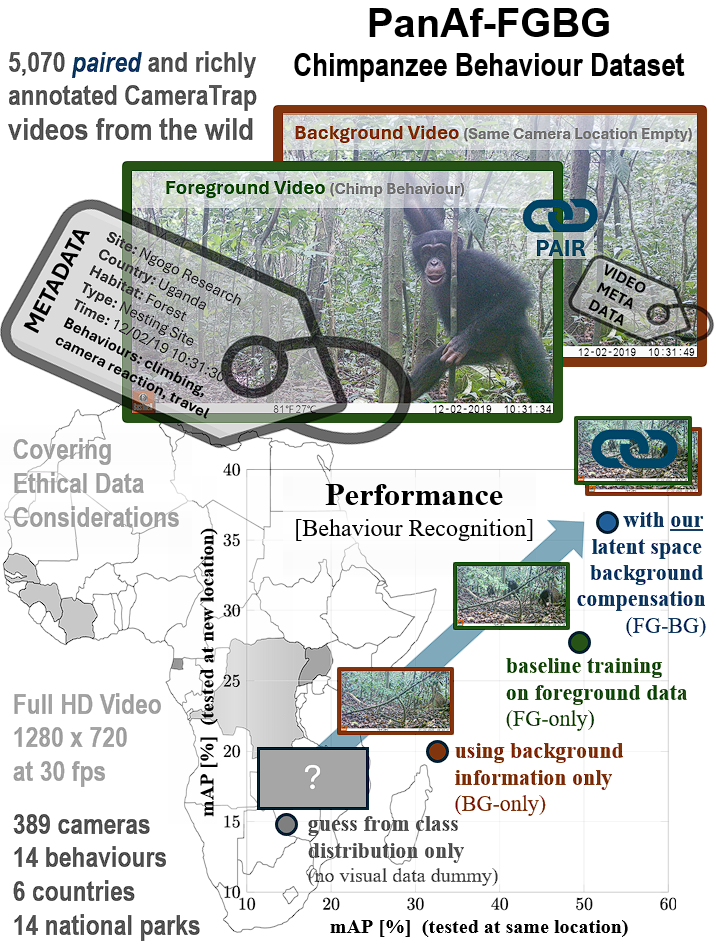}
\caption{\textmd{\textbf{{Conceptual Overview}.} The PanAf-FGBG dataset comprises $>$20 hours of paired and richly annotated foreground-background camera trap videos of wild chimpanzees. The dataset unlocks systematic analyses of the impact of background information on wildlife behaviour recognition. We provide baselines, quantify the background impact on performance, and demonstrate that utilising background information in latent space (see blue data point) can significantly improve recognition performance in this challenging domain.}}
\label{fig:overview}
\end{figure}

\section{Introduction}
\label{sec:intro}




Deep learning models have shown impressive results on action recognition tasks~\cite{zhu2020comprehensive}, particularly when both training and test data share the same distribution. However, their performance declines significantly on out-of-distribution~(OOD) data~\cite{koh2021wilds,d2022underspecification}. In practice, distribution shifts are common in nearly \textit{all} settings~\cite{quinonero2022dataset}, but are particularly unforgiving when computer vision is applied to uncontrolled, complex or natural environments~\cite{koh2021wilds}.

In the human action recognition domain, poor OOD performance is well-known to originate from shortcut learning~\cite{geirhos2020shortcut} of static cues or backgrounds~\cite{li2019repair, ilic2022appearance,li2023mitigating}. This leads models to rely heavily on background information instead of the behaviour of interest~solely~\cite{burghardt2006analysing}, and several studies have shown that the \textit{majority} of action recognition performance in this domain can be attributed to background learning~\cite{vu2014predicting, he2016human, chung2022enabling}. Despite the known impact of backgrounds on action recognition and a critical need to advance model performance in wildlife applications~\cite{tuia2022perspectives,kuhl2013animal,pollock2025harnessing}, reliance on static cues for animal behaviour recognition has received limited attention. This is largely due to a lack of datasets explicitly designed to study its effects.

In response, we introduce the PanAf-FGBG dataset, which comprises real-world foreground-background video pairs (see Fig.~\ref{fig:overview}) extracted from six African countries, 14 national parks, and over 350 individual camera locations across many habitats and location types~(see Fig.~\ref{fig:habitat_dist}). Here, we define foreground videos as those containing a chimpanzee, and background videos as those that do not. This allows us to address a key limitation in existing studies, where the background’s role is typically assessed by creating background-only versions of datasets through synthetic modifications (e.g., masking or in-painting) that remove the actor~\cite{hendricks2018women,choi2019can,chung2022enabling}. While such synthetic inputs often serve as useful proxies, the actual effect of synthesis on training is largely unknown~\cite{shankar2021image,moayeri2022comprehensive}.

In contrast, our dataset introduces paired camera locations and enables a truly realistic disentangling of background effects under in- and out-of-distribution conditions: \textit{disjoint} experiments with mutually exclusive camera locations for training and test splits, and \textit{overlapping} experiments with shared locations. This focus is particularly important in the wildlife domain, where the transferability and generalisation of camera trap analysis to new locations is critical for effective operation. While studies have shown the over-reliance of action recognition models on backgrounds~\cite{marszalek2009actions,vu2014predicting,he2016human}, we find that the effect of \textit{background duration}  (i.e., the count of background frames within foreground videos with no actor) is largely unexplored. In camera trap video footage like in many other CCTV or triggered camera settings, actors may only appear momentarily, leaving behind footage of \textit{only} the background~\cite{miao2021iterative, cunha2021filtering} within foreground videos. Since this effect alters learning given video-level behaviour annotations, we quantify its effect on several action recognition architectures.

Our contributions are as follows:
\begin{enumerate*}[label=(\roman*)] \item we release PanAf-FGBG with its rich annotations and meta-data, providing a large wildlife dataset suitable for explicitly investigating the impact of backgrounds on behaviour recognition; \item we show models trained on background-only videos achieve strong animal behaviour recognition performance 
($\sim$~65\% of baseline performance), corroborating findings in the human domain; \item we conduct the first evaluation of background duration in videos, revealing its impact on action recognition models and highlighting key differences between convolutional and attention-based networks; and \item we introduce a simple yet powerful latent-space background neutralisation that yields notable performance gains, particularly in OOD scenarios for  convolutional (+5.42\% mAP) and transformer-based~approaches~(+3.75\% mAP).\end{enumerate*}



\section{Related Work}
\label{sec:related_work}


\textbf{Animal Behaviour Datasets}. 
In recent years, several high-impact animal behaviour datasets have been 
introduced to the CV community, often accompanied by new methods~\cite{sakib2020visual, bain2021automated, ng2022animal, chen2023mammalnet,liu2023lote, brookes2023triple, kholiavchenko2024deep, brookes2024panaf20k, brookes2024chimpvlm}. Although they all centre on animal behaviour understanding, they each have a different focus. The large-scale datasets Animal Kingdom~\cite{ng2022animal} and MammalNet~\cite{chen2023mammalnet} concentrate on behavioural understanding across species, each including videos of $\sim$~800 different species. They provide atomic action and high-level behaviour annotations, respectively, and support, in addition to action recognition, the related tasks of action localisation~\cite{ng2022animal,chen2023mammalnet} and pose estimation~\cite{ng2022animal}. The smaller ChimpAct dataset~\cite{ma2023chimpact} provides longitudinal data that follows a zoo-housed group of 20 chimpanzees with a focus on a single male juvenile. It also contains key-point pose information, and fine-grained spatio-temporal behaviour labels. While valuable, these datasets lack in-situ footage from ecological sensors like camera traps and drones, meaning animals are not observed in their natural environment. This may result in unrepresentative behaviour distributions and limit their suitability for studying wild behaviour~\cite{clark2011great,chappell2022role}.


In contrast, KABR~\cite{kholiavchenko2024deep} and BaboonLand~\cite{duporge2024baboonland} comprise drone footage of free-living Kenyan wildlife, with fine-grained spatio-temporal behaviour annotations, focussing on individual and group-level behaviour, respectively. The LoTE dataset~\cite{liu2023lote} provides camera trap video footage of 11 endangered species from a 12-year longitudinal study; it is accompanied by environmental information (i.e., weather conditions and habitat) and highlights the importance of temporal and spatial continuity for conservation efforts. The dataset most comparable to our own is the PanAf20k dataset~\cite{brookes2024panaf20k} and its related PanAf500-based works~\cite{yang2019great,sakib2020visual,yang2023dynamic}. PanAf20k comprises camera trap videos of great apes accompanied by multi-label behaviour annotations.


Our dataset orthogonally complements this growing body of work dedicated to understanding animal behaviour by providing the first dataset specifically designed to evaluate backgrounds and out-of-distribution generalisation for behaviour recognition. It is unique in providing foreground-background pairs, and its configurations for in- and out-of-distribution evaluation are defined identically to the well-known iWildCam~~\cite{beery2021iwildcam} and Wilds dataset~\cite{koh2021wilds} for species classification. Similar to LotE~\cite{liu2023lote}, we provide a detailed description of the background, however we also include a unique camera ID (i.e., geospatial location) to allow footage taken from the same camera to be linked. Similar to~\cite{brookes2024panaf20k}, our data is gathered from 14 national parks, spanning 6 different countries, although unlike ours they do not include metadata to allow videos to be linked to specific countries, research sites, locations, or habitats. Similarly, it is more diverse than~\cite{liu2023lote} with its footage from a single national park.



\textbf{Background Impact}. The effect of backgrounds or static bias is well-studied within computer vision~\cite{zhang2022suppressing,li2023mitigating}, where it has shown to be \textit{necessary} for models to perform well~\cite{xiao2020noise}. Action recognition models are particularly vulnerable to shortcut learning. These models often rely on correlated, but distinct information beyond the intended action, such as object~\cite{gupta2007objects, zhou2023can}, scene~\cite{marszalek2009actions}, and actor biases~\cite{li2018resound}. For example, several studies have reported that strong performance on action recognition tasks can be achieved \textit{without} the actor~\cite{marszalek2009actions,vu2014predicting,he2016human} and that approximately 70\% of model performance is attributable to static cues (i.e., scene or background)~\cite{chung2022enabling}. This results in biased representations and ultimately impairs out-of-distribution generalization performance~\cite{li2019repair}.

\textbf{Background Bias Datasets}. The effect of the background has often been modelled using
synthetically generated data that removes either the foreground or the background.~\cite{girdhar2019cater, choi2019can, chung2022enabling, li2023mitigating}. For instance, Ilic et al.~\cite{ilic2022appearance} present a synthetic version of the UCF101 dataset, animating spatial noise using image motion extracted by an optical flow estimator, Li et al.~\cite{li2023mitigating} replace the backgrounds of IID HMDB51 with sinusoidal stripe images, and Chung et al~\cite{chung2022enabling} adopt an in-painting model to remove actors from the video. Datasets that include fine-grained spatio-temporal action annotations also remove static cues implicitly by isolating the region of the video where the action is occurring.



\textbf{Mitigating Background Bias}. Prominent attempts to mitigate this bias include dataset resampling methods, which down-weight biased samples~\cite{li2018resound,li2019repair}, adversarial training regimes that rely on scene labels and human masks~\cite{choi2019can}, and augmentation techniques that mix or swap the backgrounds of different videos during training~\cite{wang2021removing, ding2022motion,li2023mitigating}. There are also several implicit approaches, such as frame selection, that aim to utilise only frames where activity is occurring~\cite{zhao2023search}. Our datatset, introduced next, instead introduces foreground-background pairing information with close proximity in time of capture, which describes the environment truly independently and in relation to the presence of animals showcasing behaviour.

\begin{figure}[!b]
\centering
\includegraphics[width=0.95\linewidth]{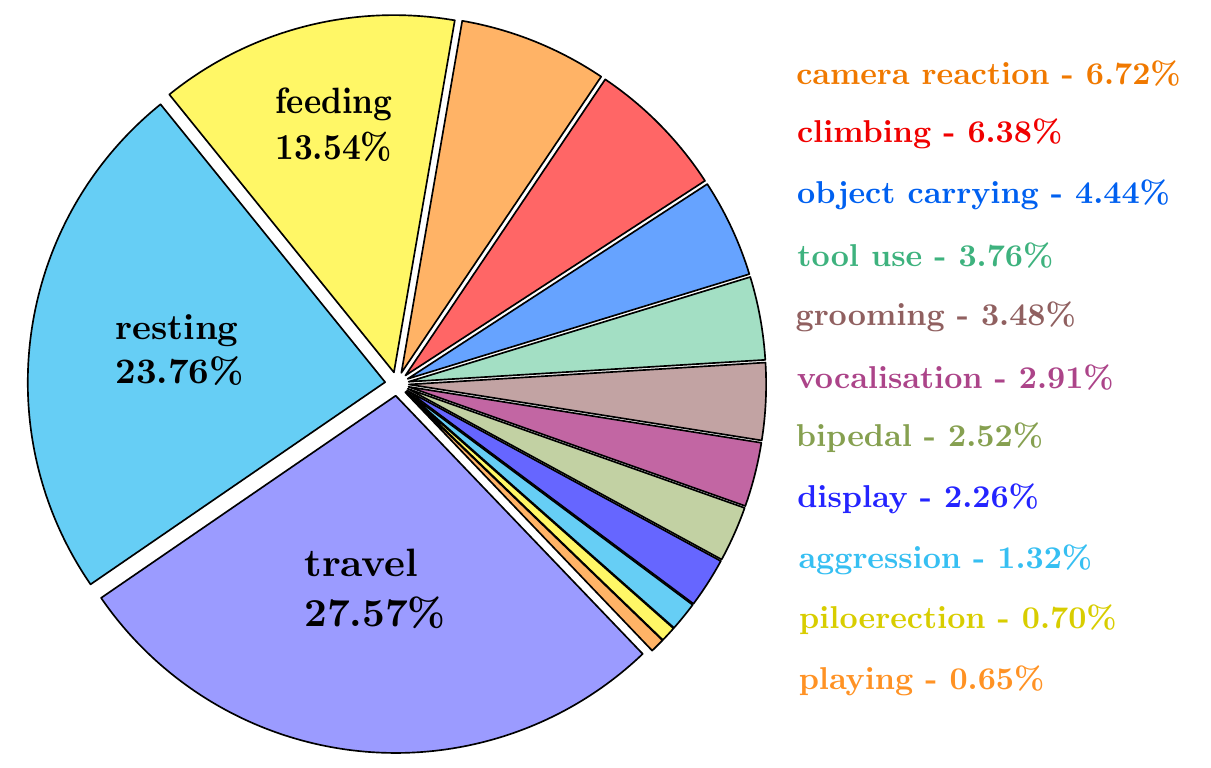}
\caption{{{\textmd{\textbf{Distribution of Behaviour}. Proportion of behaviours in the dataset where smaller segments are colour coordinated.}}}}
\label{fig:class_dist}
\end{figure}

\section{Dataset}\vspace{-4pt}
\label{sec:dataset}

\begin{figure}[!h]
\centering
\includegraphics[width=1.0\linewidth]
{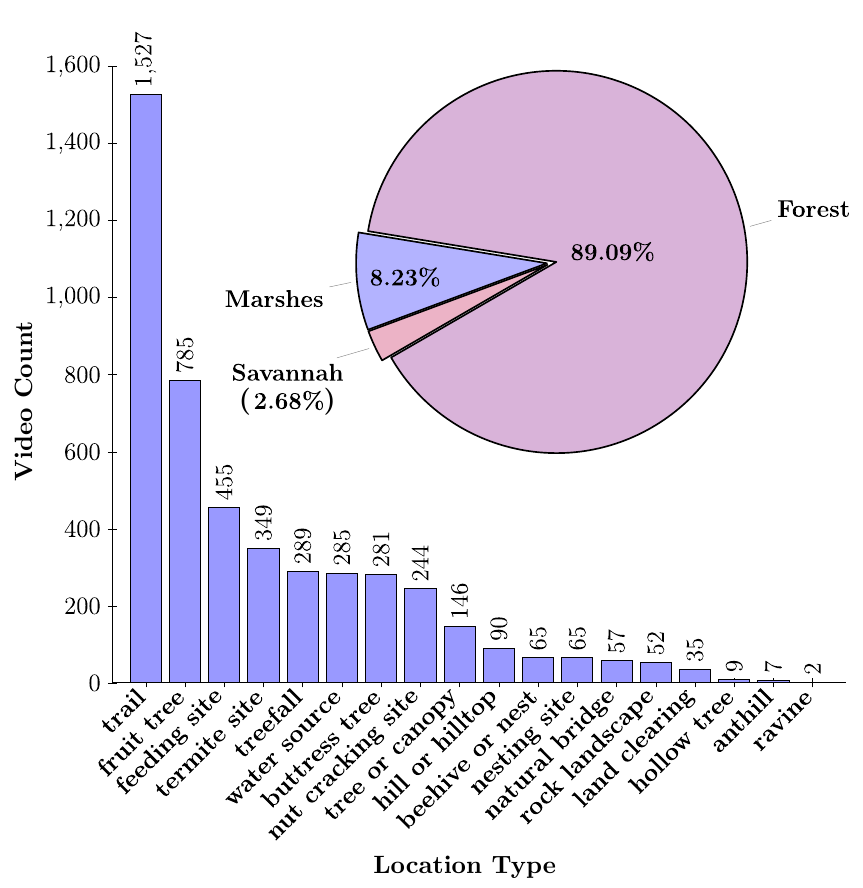}
\caption{{\textmd{\textbf{Distribution of Habitat and Location Type}. The pie chart gives the proportion of videos taken at each of the three main habitats: forest, marshes and savannah.The histogram shows the total number of videos extracted from each location type.}}}
\label{fig:habitat_dist}
\end{figure}

\begin{figure*}[!h]
\centering
\includegraphics[width=1.0\linewidth]{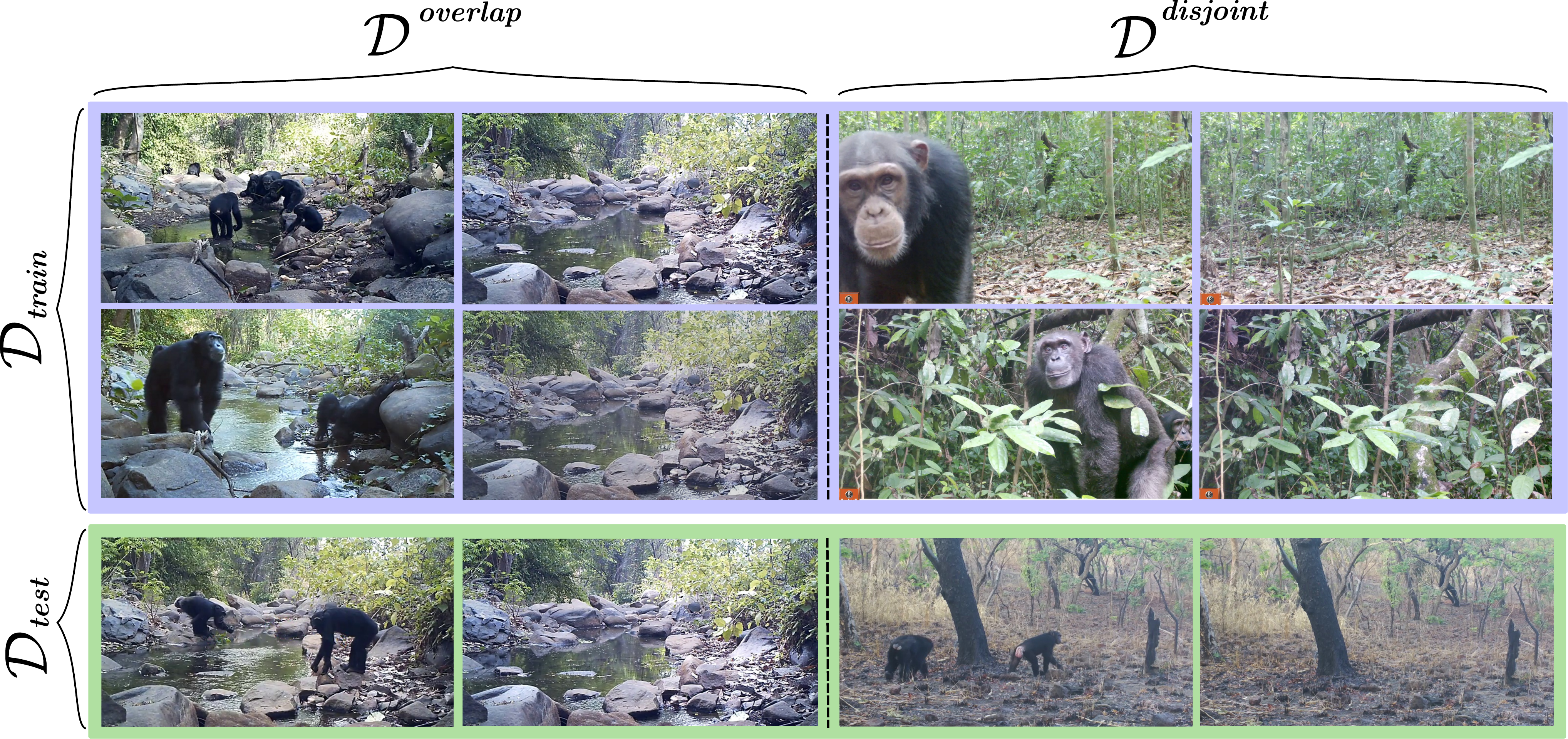}
\caption{{\textmd{\textbf{Overlapping \& Disjoint Dataset Configurations}. Six foreground-background video pairs are shown to visualise overlapping (left) and disjoint (right) dataset configurations. On the left, train and test videos taken from the same camera are selected, emphasizing that the locations are not mutually exclusive, with backgrounds containing several cues that are highly correlated with the displayed behaviours (i.e., algae fishing, a form of tool use). Train and test videos on the right are extracted from mutually exclusive or \textit{disjoint} camera locations. They are extracted from forest and savannah, respectively, and highlight the challenge of generalising to new locations.}}}
\label{fig:dataset_overlap_disjoint}
\end{figure*}



{\bf Dataset Overview and Statistics.} PanAf-FGBG comprises footage gathered under ethical oversight as part of the PanAf Programme: The Cultured Chimpanzee~\cite{PanAfDataCollection}. It contains 21 hours of camera trap footage of individual chimpanzees in tropical Africa. Its footage is collected from 389 
individual camera locations across 14 national parks in 6 African countries. In total, we provide 5,070 video pairs, each 15 seconds in duration. All videos are accompanied by a multi-label behaviour annotation and scene metadata, plus the unique feature of foreground-background video pairing~(see Fig.~\ref{fig:overview}). Class and habitat distributions of the data are given in Fig.~\ref{fig:class_dist} and Fig.~\ref{fig:habitat_dist}, respectively. Our dataset follows a long-tailed distribution~\cite{liu2019large,perrett2023use} common in animal behaviour datasets~\cite{ng2022animal,chen2023mammalnet,liu2023lote, ma2023chimpact, brookes2024panaf20k}. The three of the most commonly occurring classes are observed in $>$60\% of videos, while the rarest are observed in fewer than $<$~3\%.


{\bf Data Collection.} The data collection strategy was designed specifically to maximise the chance of filming the terrestrial activity of apes and we provide only a brief description here (for full details we refer the reader to 
\cite{PanAfDataCollection}). Data was collected from 14 different research sites, where an average of 29 movement-triggered Bushnell cameras were installed per site. Grids comprising 20 to 96 $1\times1$ km cells were established and one camera was installed per grid cell. Footage was recorded at 30 frames-per-second and a resolution of $1280\times720$. The cameras were visited every 1-3 months for maintenance and to download the recorded footage throughout the study periods.

{\bf Ethical Considerations.} To protect the location of the apes from threats (e.g., poaching) we use monikers for the names of the research sites and assign a unique camera ID to each video, hiding the geospatial location of the cameras.




{\bf Data Annotation.} The behavioural annotations were provided by users on the community science platform Chimp\&See~\cite{arandjelovic2016chimp}. Annotators were presented with a choice of behaviour classification categories, chosen specifically by experts for their ecological importance. Behaviours not listed in the classification categories were also permitted. These annotations were then extracted and expertly grouped into 14 co-occurring classes, which form the multi-label behavioural annotations presented here. The annotations follow a multi-hot binary format that indicate the presence of one or more behaviours. It should also be noted that behaviours are not assigned to individual apes or temporally localised within each video. To ensure annotation quality and consistency a video was only deemed to be analysed when either three volunteers marked the video as blank, or unanimous agreement between seven volunteers was observed, or 15 volunteers annotated the video. Habitat type was noted during camera setup.



{\bf Foreground-Background Pair Selection.} Pairs are formed by selecting the temporally closest available foreground and background videos to ensure similarity. Complete standardisation statistics can be found in Sup. Mat.

{\bf Overlapping \& Disjoint Configurations.} We present both a \textit{disjoint} view of dataset, where camera locations in the training and test splits are mutually exclusive, denoted as \( \mathcal{D}^{disjoint} = \{d_i \mid d_i \cap d_j = \emptyset, \; \forall i \neq j\} \) following~\cite{koh2021wilds}. A second, \textit{overlapping} view is also provided, where camera locations are shared between training and test splits, represented as \( \mathcal{D}^{overlap} = \{d_i \mid d_i \cap d_j \neq \emptyset \; \text{for some} \; i \neq j\} \). Note that~\( d_i \) refer to sets of camera locations associated with videos in each split. Fig.~\ref{fig:dataset_overlap_disjoint} provides an overview of both configurations. We ensure that the class distribution in each configuration remains approximately the same for full comparability. Unique camera identifiers are available for each video to allow researchers interested in this problem to create additional configurations, should they wish to.




\section{Preliminaries}

We train multi-label classification models on both views of the dataset: \( \mathcal{D}^{\text{overlap}} \) and \( \mathcal{D}^{\text{disjoint}} \) using various combinations of foreground (\( \mathcal{F} \)), background ( \( \mathcal{B} \)) and synthetic background (\( \tilde{\mathcal{B}}\)) videos. The latter were created by generating a segmentation mask using SAM2~\cite{ravi2024sam} followed by mean pixel value filling, similar to~\cite{hendricks2018women,choi2019can,chung2022enabling}. Both views comprise foreground-background video pairs \( \{ (v^{\mathcal{F}}_{i}, v^{\mathcal{B}}_{i}) \}_{i=1}^N \), where \( v^{\mathcal{F}}_{i} \in \mathcal{F} \) and \( v^{\mathcal{B}}_{i} \in \mathcal{B} \), with \( N \) representing the total number of video pairs. Each foreground video \( v^{\mathcal{F}}_{i} \) is associated with a video-level label vector \( y_{i} = (y_{i1}, y_{i2}, \dots, y_{iC}) \), where each \( y_{ij} \in \{0, 1\} \) indicates the presence (\(1\)) or absence (\(0\)) of the \( j \)-th class in video \( i \), and \( C \) is the total number of classes.

\section{Experiments}
\label{sec:experiments}



\subsection{Setup}
\label{subsec:setup}

All experiments were performed using the ResNet-50~\cite{he2016deep} and MViT-V2~\cite{li2022mvitv2} architectures, which were initialised with feature extractors pre-trained on Kinetics-400~\cite{kay2017kinetics}. For ResNet-50, both 2D and 3D convolution versions are tested. After initial experiments to ensure optimal optimiser-network pairings for the task, ResNet and MViT-V2 models were fine-tuned for 300 epochs using the SGD~\cite{robbins1951stochastic} and AdamW~\cite{loshchilov2017decoupled} optimisers, respectively. Both models followed a linear warm-up schedule with subsequent cosine annealing. For ResNet models, the learning rate was increased from $1\times10^{-5}$ to $1\times10^{-4}$ over 20 epochs, while it was increased from $1\times10^{-6}$ to $1\times10^{-4}$ over 30 epochs for MViT-V2 models. In both cases, momentum of 0.9 was utilised. All computation was performed on 4$\times$NVIDIA~H200 GPUs using a distributed batch size of 336. During training, video tensors with a spatio-temporal resolution of $16\times256\times256$ were employed. Frames were sampled uniformly at a rate of $1/24$, avoiding dense sampling strategies which are shown to exacerbate shortcut learning in action recognition models~\cite{chung2022enabling}. For MViT-V2 models, a patch size of $16\times16$ pixels was utilised. Evaluation was performed using micro and mean Average Precision and are referred to as uAP and mAP, respectively.


\subsection{Background Reliance}
\label{subsec:bg_reliance}

\textbf{Experimental Overview}. First, we examined the reliance of animal behaviour recognition on the background by training  model architectures on the foreground~${\mathcal{F}}$, background~${\mathcal{B}}$, and synthetic background videos~$\tilde{\mathcal{B}}$ only. We refer to models trained on ${\mathcal{F}}$ and ${\mathcal{B}}$ (or $\tilde{\mathcal{B}}$) as FG-only and BG-only models, respectively, as shown earlier in Fig.~\ref{fig:overview}. To separate background information from class-distribution intrinsic information, a dummy classifier with no access to visual information was also evaluated. It would effectively return predictions based on the class distribution information alone. Since our dataset consists of unique foreground-background pairs, it was possible to train BG-only models by retaining the original labels associated with the foreground video of the pairing. In all cases, models were tested on~${\mathcal{F}}$ for comparability. Note that this makes the task more difficult for BG-only models, since animals are only observed at test time and not during training. All experiments are conducted on both \(D^{overlap}\) and \(D^{disjoint}\).

\textbf{Dual-Stream Fusion Model ($\mathcal{F}+\mathcal{B}$)}. To clarify whether BG-only models and FG-only models, each trained separately, can be fused late to orthogonally contribute towards performance we also implemented a dual-stream model. To do this, the independently optimised weights of the FG-only and BG-only models trained for 300 epochs were loaded. During training the weights were frozen and video pairs were processed to extract late features \(z^{F}\) and \(z^{B}\). For ResNet models, features were extracted from the last convolutional layer and in MViT-V2 models an average of the tokens from the last hidden state were taken. Late features were then concatenated \(\tilde{z} = [z^{F}, z^{B}]\) and final class predictions were produced by a trainable multi-layer perceptron.



\begin{table}[t]
\footnotesize
    \centering
    \begin{tabular}{llcccccc}
        \toprule
        & \multirow{2}{*}{Model} & \multirow{2}{*}{Data} & \multicolumn{2}{c}{$\mathcal{D}^{overlap}$} & \multicolumn{2}{c}{$\mathcal{D}^{disjoint}$} \\
        \cmidrule(l){4-5}\cmidrule(l){6-7}
        & & & uAP & mAP & uAP & mAP \\
        \midrule
        \rowcolor{gray!15}
        1 & Dummy & - & 19.15 & 14.73 & 19.61 & 14.87\\
        \hdashline
        2 & \multirow{4}{*}{2D R50} & $\mathcal{F}$ & 64.15 & 41.28 & 44.67 & 18.08\\
        3 & & $\mathcal{B}$ & 57.13 & 30.03 & 45.80 & 18.27\\
        4 & & $\tilde{\mathcal{B}}$ & 57.87 & 30.82 & 48.28 & 20.39\\
        5 & & $\mathcal{F}+\mathcal{B}$ & 67.74 & 44.99 & 45.28 & 18.84\\
        \rowcolor{cyan!15}
        6 & & & \textit{0.89} & \textit{0.74} & \textit{1.02} & \textit{0.98}\\
        \hdashline
        7 & \multirow{4}{*}{3D R50} & $\mathcal{F}$ & 72.79 & 49.32 & 57.74 & 30.89\\
        8 & & $\mathcal{B}$ & 58.92 & 31.72 & 48.40 & 21.38\\
        9 & & $\tilde{\mathcal{B}}$ & 59.60 & 32.86 & 49.29 & 21.64\\
        10 & & $\mathcal{F}+\mathcal{B}$ & \textbf{76.41} & \textbf{53.57} & 65.66 & \textbf{39.17}\\
        \rowcolor{cyan!15}
        11 & & & \textit{0.80} & \textit{0.66} & \textit{0.83} & \textit{0.69}\\
        \hdashline
        12 & \multirow{4}{*}{MViT-V2} & $\mathcal{F}$ & \underline{75.44} & \underline{51.54} & \underline{\textbf{70.50}} & \underline{35.29}\\
        13 & & $\mathcal{B}$ & 61.90 & 32.74 & 50.03 & 24.00\\
        14 & & $\tilde{\mathcal{B}}$ & 62.15 & 33.58 & 52.13 & 24.77\\
        15 & & $\mathcal{F}+\mathcal{B}$ & 72.82 & 53.49 & 68.00 & 38.09\\
        \rowcolor{cyan!15}
        16 & & & \textit{0.82} & \textit{0.65} & \textit{0.70} & \textit{0.68}\\
        \bottomrule
    \end{tabular}
    \caption{\textbf{Quantifying Background Reliance}. \textmd{Performance comparison of models trained on \(\mathcal{F}\), \(\mathcal{B}\), and \(\tilde{\mathcal{B}}\) videos. Dual-stream models are indicated by $\mathcal{F}+\mathcal{B}$. The absolute highest performance is indicated in \textbf{bold} whereas best performance using only a single-stream is \underline{underlined}. The performance based on class distribution only is given by a dummy classifier (\colorbox{gray!15}{see row 1}). The background-to-foreground performance ratio (\(\frac{AP^B}{AP^F}\)) is also provided \colorbox{cyan!10}{(see rows 6, 11, and 16)}. Performance is reported on \(\mathcal{D}^{overlap}\) and \(\mathcal{D}^{disjoint}\) where Fig.~\ref{fig:overview} visualises an mAP scatter plot between the two for 3D R50.}}
    \label{tab:bg_reliance}
\end{table}

\textbf{Result 1 -- Backgrounds are a Predictor of Animal Behaviour.} Table~\ref{tab:bg_reliance} shows that BG-only models achieve strong performance when compared to their respective FG-only counterparts. To illustrate this, the ratio of background-to-foreground performance \(\frac{AP^B}{AP^F}\) published in~\cite{chung2022enabling}, showing how close a model can get to FG performance with just BG cues, is reported for each architecture in rows 6, 11, and 16. Background-to-foreground performance ratios are consistently high across all models, never dropping below 0.70 and 0.65 for uAP and mAP, respectively, on either dataset. This indicates that much of the behaviour recognition performance is \textit{indeed} achievable by just utilisation of background information - the environment is a strong predictor of behaviour, even out-of-distribution. Thus, features of the environment are already behaviour predictors~(trees for climbing, paths for travel, termite mound for feeding etc.). This result is further compounded by the fact that BG-only models outperform the non-visual classifier~(row 1) by a significant margin, particularly on $\mathcal{D}^{overlap}$. On $\mathcal{D}^{disjoint}$ this observation still holds, although by somewhat smaller margins. Note that for the weakest model (2D R50) background mAP is even on-par or above the foreground baseline~(see rows 2-3 in the final column). Together, these results illustrate powerfully, that across architectures background information alone is employable as a strong animal behaviour predictor.

\textbf{Result 2 -- Multi-scale Vision Transformers rely less on backgrounds.} On the overlapping dataset, the 3D-R50 and MViT-V2 models~(rows 7-16) display significantly better performance when compared with the 2D-R50~(rows 2-6), and perform similarly to each other with respect to uAP and mAP. However, the MViT-V2 model widely outperforms the 3D R50 on the disjoint dataset, suggesting that it does not rely as heavily on background features for performance. In support of this, while all models experience a large decrease in performance when moving from the overlapping to the disjoint dataset (generalisation gap), the smallest impact is on the MViT-V2. These differences between architectures are highlighted most clearly by observing uAP on the disjoint dataset. Here, the background only ratio for the MViT-V2 model is $\sim$~30\% and $\sim$~13\% lower than the 2D-R50 and 3D-R50, respectively. MViT-V2 achieves the best overall performance and exhibits the lowest background-to-foreground performance ratio on both dataset configurations, corroborating similar findings from the still image domain that show transformers are more invariant to changes in the background~\cite{moayeri2022comprehensive}.


\textbf{Result 3 - Analysing Background Information Contributions.} The dual 3D-R50 model achieves significant performance gains over its FG-only baseline, particularly on $\mathcal{D}^{disjoint}$, where increases of 7.92\% and 8.28\% in uAP and mAP, respectively, can be observed. Smaller, but still significant improvements are also observed on $\mathcal{D}^{overlap}$ (+3.62\% uAP and +4.25\% mAP). Interestingly, the performance of the dual-stream 3D-R50 exceeds that of MViT-v2, including its dual-stream counterpart, on nearly all metrics except for disjoint uAP. The dual-stream MViT-V2 model, on the other hand, shows improved mAP but a lower uAP when compared with the baseline model, a trend consistent across both dataset configurations.
This analysis indicates that BG-only performance in the wildlife domain is \textit{not} a subset of the FG-only model -- which of course also has access to background information. Instead, explicitly learned (and frozen) background features can aid the FG-only baseline in models such as ResNet via late fusion. Finally, all BG-only models trained on $\tilde{\mathcal{B}}$ seem to outperform those trained on $\mathcal{B}$. However, we will show later in Sec.~\ref{subsec:bg_mitigation} that non-synthetic background videos dominate synthetic counterparts when explicitly utilised for performance improvements above the foreground baselines.

\subsection{Background Duration}
\label{subsec:temporal_negtivity}

\begin{figure}[h]
\centering
\includegraphics[width=1.0\linewidth]{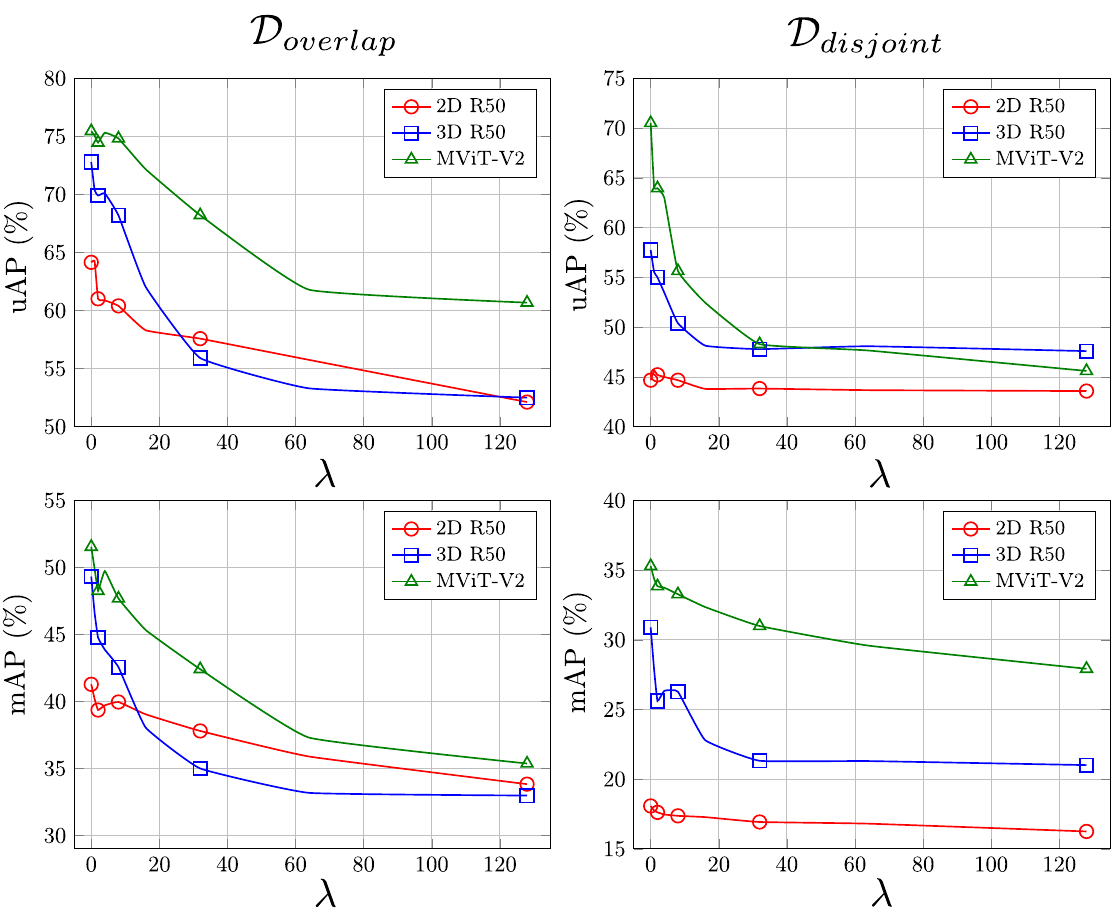}
\caption{\textmd{\textbf{Effect of Increasing Background Duration on Performance}. Comparison of 2D R50, 3D R50 and MViT-V2 model performance  as a function of the simulated background duration ($\lambda$). The 3D-R50 is more sensitive to increasing background durations than MViT-V2. MViT-V2 is relatively robust to increasing background durations on the overlapping dataset, although it becomes highly sensitive when evaluated on OOD data.}}
\label{fig:background_duration}
\end{figure}

\textbf{Experimental Overview.} Noting that foreground videos may contain some frames with no animals, effectively supplying background only information for these frames, we aim at quantifying the effect of this on model performance~(see Fig.~\ref{fig:background_duration}). Practically, this is achieved by concatenating increasing numbers of background frames to our foreground videos. To do this, we sampled a set of \( \lambda T \) frames from \( v^{B} \), where \( \lambda \) is a predefined factor controlling the proportion of background frames used. The sampled background frames are denoted as \( V^{\mathcal{B}} = \{ V_{i_0}^{\mathcal{B}}, V_{i_1}^{\mathcal{B}}, \dots, V_{i_{\lambda T - 1}}^{\mathcal{B}} \} \), with each index \( i_j \) selected from \( \{ 0, \dots, T - 1 \} \). The foreground frames \( V^{\mathcal{F}} = \{ F_0^{\mathcal{F}}, F_1^{\mathcal{F}}, \dots, F_{T-1}^{\mathcal{F}} \} \) were then concatenated with \( V^{\mathcal{B}} \), resulting in a combined video sequence \( V^{\text{concat}} = \{ F_0^{\mathcal{F}}, F_1^{\mathcal{F}}, \dots, F_{T-1}^{\mathcal{F}}, F_{i_0}^{\mathcal{B}}, F_{i_1}^{\mathcal{B}}, \dots, F_{i_{\lambda T - 1}}^{\mathcal{B}} \} \) of length \( T + \lambda T \). From this concatenated sequence, a starting frame \( F_{\text{start}} \) was sampled at random from the indices \( \{0, \dots, \lambda T\} \), ensuring that \( T \) consecutive frames could be sampled without exceeding the sequence bounds. The final sampled frame sequence \( S = \{ F_{\text{start}}, F_{\text{start} + 1}, \dots, F_{\text{start} + T - 1} \} \) was then used as input for training while retaining the original label \( y \) from the foreground video \( v^{\mathcal{F}} \). Background frames are appended because motion-triggered cameras are more likely to capture activity first, resulting in a more natural ordering.

\textbf{Analysing the Continuum Between Foreground and Background Videos.} Fig.~\ref{fig:background_duration} shows the performance comparison between the 2D R50, 3D R50 and MViT-V2 models as background duration increases in videos ($\lambda$). As expected, Fig.~\ref{fig:background_duration} confirms that as $\lambda$ increases model performance declines towards background model performance (see Tab~\ref{tab:bg_reliance}). On $\mathcal{D}^{overlap}$, the 3D-R50 and MViT-V2 models exhibit a sharp decline before beginning to converge. This occurs at $\lambda=32$ and $\lambda=64$ for the 3D-R50 and MViT-V2 models, respectively. Additionally, as shown by the interval between $\lambda=0$ and $\lambda=32$, the uAP and mAP of 3D-R50 falls by 16.88\% and 14.32\%, respectively, while falling by only 7.23\% and 9.13\% for the MViT-V2 model. Earlier convergence to near BG-only performance and a more rapid rate of decline, show that 3D-R50 is more sensitive to increases in background durations than MViT-V2. Despite starting from lower performance, the 2D R50 model exhibits a more gradual, near-linear decline, indicating less sensitivity to changes in $\lambda$. While the behaviour of the 2D-R50 and 3D-R50 models remain relatively consistent on $\mathcal{D}^{disjoint}$, the behaviour of MViT-V2 changes significantly. Its uAP declines more steeply, falling from 70.05\% to 48.29\% (a decrease of 22.21\%) between $\lambda=0$ and $\lambda=32$, where its performance converges. It is also more sensitive to smaller increases in background durations, falling from 70.50\% to 64.08\% (a decrease of 6.42\%) between $\lambda=0$ and $\lambda=1.0$, where a decrease of only 0.38\% is experienced on $\mathcal{D}^{overlap}$ over the same interval. At higher values of $\lambda$~(i.e., $>40$) the performance drops below that of the 3D-R50 (although only once the background-only performance has been surpassed). Although uAP falls more steeply, mAP declines more slowly and performance above that of the BG-only model is maintained even at the highest values of $\lambda$. This suggests that, although the MViT-V2 achieves higher baseline performance on the disjoint dataset, it is more sensitive to increases in background frames when evaluating uAP; a scenario where the model's advantage is drowned out by background information.

\begin{figure}[b]
\centering
\includegraphics[width=1.0\linewidth]{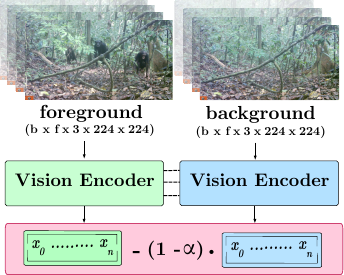}
\caption{{\textmd{\textbf{Latent Space Background Compensation}. The proposed operation uses \(\alpha\) to perform a weighted subtraction of background features (computed via the dataset pairing of a real-world, background video) from foreground features in latent space. The dotted line between the encoders indicates shared weights (i.e., they are the same model).}}}
\label{fig:arch_alone}
\end{figure}

\subsection{Background Bias Mitigation}
\label{subsec:bg_mitigation}

{\textbf{Augmenting Performance via Explicit Background Neutralisation.} In this final experimental section, we justify the need for paired, real-world background samples, as uniquely provided by our dataset. We show that these sample pairings are able to significantly improve model performance.
To show this, we analyse the impact of background subtraction operations in both input and embedding spaces across each of the model architectures and dataset configurations. Two background-subtracted views of our dataset were created for input space experiments. This was done by performing pixel-wise subtraction of $\mathcal{B}$ and $\tilde{\mathcal{B}}$ from $\mathcal{F}$. To perform algebraic background subtraction in latent space, we extracted the feature vectors of the foreground video pairs. As described in Sec.~\ref{subsec:bg_reliance}, we achieve this by processing the video tensors up to the last convolutional layer and averaging the tokens from the last hidden state in the ResNet and MViT-V2 models, respectively. The background-subtracted embedding is then computed as $z^{\mathcal{F}-\mathcal{B}} = z^{\mathcal{F}} - (1 - \alpha) \cdot z^{\mathcal{B}}$ where \( \alpha \) is a coefficient used to control the relative magnitude of the algebraic subtraction operation as schematically shown in Fig.~\ref{fig:arch_alone}. In practice, the modulation parameter \( \alpha \in [0, 1] \) was ablated to increase linearly or exponentially over the course of training and results are reported for both. In all cases \(\alpha\) was initialized at zero and increased to one by the end of training.




\begin{table}[!htbp]
\footnotesize
    \centering
    \begin{tabular}{lccccc}
        \toprule
        \multirow{2}{*}{Model} & \multirow{2}{*}{Subtraction}  & \multicolumn{2}{c}{$\mathcal{D}^{overlap}$} & \multicolumn{2}{c}{$\mathcal{D}^{disjoint}$} \\
        \cmidrule(l){3-4}\cmidrule(l){5-6}
        & & uAP & mAP & uAP & mAP \\
        \midrule
   
         &Baseline & 64.15 & 41.28 &44.67 & 18.08\\
        &$\mathcal{F}+\mathcal{B}$ & \underline{67.74} & \underline{44.99} & 45.28 & 18.84\\      
        &\cellcolor{input_space!20!white}\(\mathcal{F}\ -\tilde{\mathcal{B}}\) & \cellcolor{input_space!20!white}65.96 &\cellcolor{input_space!20!white}42.12 &\cellcolor{input_space!20!white}\textbf{54.53} & \cellcolor{input_space!20!white}\textbf{26.65}\\
         &\cellcolor{input_space!20!white}\(\mathcal{F}\ - \mathcal{B}\) & \cellcolor{input_space!20!white}\textbf{72.21}& \cellcolor{input_space!20!white}\textbf{47.12} & \cellcolor{input_space!20!white}\underline{53.73}& \cellcolor{input_space!20!white}\underline{25.84}\\
         &\cellcolor{embedding_space!20!white}None & \cellcolor{embedding_space!20!white}56.21 & \cellcolor{embedding_space!20!white}38.84 &\cellcolor{embedding_space!20!white} 27.24 &\cellcolor{embedding_space!20!white} 17.94\\
         &\cellcolor{embedding_space!20!white}Linear & \cellcolor{embedding_space!20!white}53.78 &\cellcolor{embedding_space!20!white}39.01 &\cellcolor{embedding_space!20!white} 29.75 & \cellcolor{embedding_space!20!white}17.17\\
         \multirow{-6}{*}{2D R50} &\cellcolor{embedding_space!20!white}Exponential &\cellcolor{embedding_space!20!white} 54.05 & \cellcolor{embedding_space!20!white}37.11 & \cellcolor{embedding_space!20!white}25.80 & \cellcolor{embedding_space!20!white}17.73\\
      
        \hdashline
          &Baseline & 72.79 & 49.32 & 57.74 & 30.89\\
       &$\mathcal{F}+\mathcal{B}$ & \underline{76.41} & \textbf{53.57} & \underline{65.66} & \textbf{39.17}\\
         &\cellcolor{input_space!20!white}\(\mathcal{F}\ -\tilde{\mathcal{B}}\) & \cellcolor{input_space!20!white}66.51 & \cellcolor{input_space!20!white}42.56 & \cellcolor{input_space!20!white}53.58 & \cellcolor{input_space!20!white}27.98\\
         &\cellcolor{input_space!20!white}\(\mathcal{F}\ - \mathcal{B}\) &\cellcolor{input_space!20!white} 71.93 &\cellcolor{input_space!20!white}46.36 & 53.92\cellcolor{input_space!20!white} & \cellcolor{input_space!20!white}25.05\\

       &         \cellcolor{embedding_space!20!white}None &          \cellcolor{embedding_space!20!white}72.30 &          \cellcolor{embedding_space!20!white}51.39 &          \cellcolor{embedding_space!20!white}63.14 &          \cellcolor{embedding_space!20!white}34.85\\
        &\cellcolor{embedding_space!20!white}Linear  &\cellcolor{embedding_space!20!white} \textbf{77.31} &\cellcolor{embedding_space!20!white}\underline{52.82}&\cellcolor{embedding_space!20!white}\textbf{66.69}& \cellcolor{embedding_space!20!white}36.31\\
         \multirow{-6}{*}{3D R50} &\cellcolor{embedding_space!20!white}Exponential &\cellcolor{embedding_space!20!white} 73.52 & \cellcolor{embedding_space!20!white}51.24 & \cellcolor{embedding_space!20!white}64.58 & \cellcolor{embedding_space!20!white}\underline{36.36}\\
        \hdashline
          &Baseline & 75.44 &51.54 & 70.50 & 35.29\\
          &$\mathcal{F}+\mathcal{B}$ & 72.82 & \underline{53.49} & 68.00 & 38.09\\
         &   \cellcolor{input_space!20!white}\(\mathcal{F}\ -\tilde{\mathcal{B}}\)
         &\cellcolor{input_space!20!white} 66.84 & \cellcolor{input_space!20!white}42.27 & \cellcolor{input_space!20!white}57.21 & \cellcolor{input_space!20!white}28.80\\
         &\cellcolor{input_space!20!white}\(\mathcal{F}\ - \mathcal{B}\)
         &\cellcolor{input_space!20!white} 74.90 &\cellcolor{input_space!20!white}49.40 & \cellcolor{input_space!20!white}61.22 & \cellcolor{input_space!20!white}31.51\\
        
         &\cellcolor{embedding_space!20!white}None &\cellcolor{embedding_space!20!white} 74.23 &\cellcolor{embedding_space!20!white} 44.11 & \cellcolor{embedding_space!20!white}68.56 &\cellcolor{embedding_space!20!white} 35.14\\
         &\cellcolor{embedding_space!20!white}Linear &\cellcolor{embedding_space!20!white} \textbf{79.32$^*$} &\cellcolor{embedding_space!20!white}\textbf{53.83$^*$} & \cellcolor{embedding_space!20!white}\textbf{71.64$^*$} &\cellcolor{embedding_space!20!white}\underline{39.04}\\
         \multirow{-6}{*}{MViT-V2} &\cellcolor{embedding_space!20!white}Exponential & \cellcolor{embedding_space!20!white}\underline{77.70} & \cellcolor{embedding_space!20!white}50.93 & \cellcolor{embedding_space!20!white}\underline{71.28} & \cellcolor{embedding_space!20!white}\textbf{39.25$^*$}\\
        \bottomrule
    \end{tabular}
    \caption{\textbf{Quantifying Performance Improvements of Background Subtraction.} \textmd{Comparison of the 2D R50, 3D R50, and MViT-V2 models on both dataset configurations. Performance is reported when training on  \(\mathcal{F}\ -\mathcal{B}\) or \(\mathcal{F}\ -\tilde{\mathcal{B}}\) videos in input (\colorbox{input_space!20}{light green}) and embedding space (\colorbox{embedding_space!20}{light blue}). \textbf{First} and \underline{second} highest scores are indicated for each model, while the highest overall scores are marked by an asterisks ($*$). Note the baseline is trained on \(\mathcal{F}\), as reported in Tab.~\ref{tab:bg_reliance}. $\mathcal{F}+\mathcal{B}$ provides an alternative baseline but is reliant on background videos at test time. Results confirm the superiority of real-world background videos over synthetic data, and of latent space neutralisation over input space background subtraction.}}
    \label{tab:quant_bg_sub}
\end{table}

\textbf{Results and Performance Improvements}. Tab.~\ref{tab:quant_bg_sub} shows that for the basic 2D-R50 model, background subtraction in input space is highly effective. When using $\mathcal{B}$, we observe substantial performance improvements over the baseline, with percentage increases of +8.06\% and 5.84\% in uAP and mAP on \( D^{overlap} \), and +9.06\% and +7.76\% in uAP and mAP on \( D^{disjoint} \). Using $\tilde{\mathcal{B}}$ also yields improvements, though these gains are notably smaller. Background subtraction in embedding space does not provide any benefit to the 2D-R50. Although input space background subtraction is effective for the 2D-R50, it does not improve performance for the 3D-R50 and MViT-V2 models. Neither original nor synthetic backgrounds provide an advantage over their respective baselines, indicating that input space subtraction may not suit spatio-temporal architectures.

Conversely, background subtraction in embedding space provides significant performance improvements for the 3D-R50 and MViT-V2 models, particularly on \( D^{disjoint} \). For the 3D-R50, embedding space subtraction with linear scheduling results in a +8.95\% and +5.42\% increase in uAP and mAP, respectively. Exponential scheduling also yields improvements over the baseline. The improvements are lower than linear scheduling with respect to uAP (+6.84\%) but marginally better for mAP (+5.47\%). MViT-V2 also benefits from embedding space subtraction on \(D^{disjoint}\): linear scheduling improves uAP and mAP by +1.14\% and +3.75\%, respectively, whereas exponential scheduling provides a +0.78\% increase in uAP and a +3.96\% increase in mAP. These findings suggest that embedding space subtraction is effective for both 3D-R50 and MViT-V2 models on \( D^{disjoint} \), with linear and exponential scheduling favouring uAP and mAP performance, respectively. Embedding space subtraction also enhances performance on \(D^{overlap}\), though to a lesser extent than on \(D^{disjoint}\). For the 3D-R50 model, applying embedding space subtraction with linear scheduling yields an improvement of 4.52\% and 3.50\% in uAP and mAP, respectively, while exponential scheduling results in smaller gains of 0.73\% and 1.92\%. The MViT-V2 model follows a similar trend: linear scheduling increases uAP and mAP by 3.88\% and 2.29\%, respecitvely, whereas exponential scheduling improves uAP by 2.26\% but slightly reduces mAP by -0.61\%. These results suggest that algebraic embedding space subtraction is beneficial on \(D^{overlap}\), although the performance gains are relatively small compared to \(D^{disjoint}\), and that linear scheduling generally provides better results than exponential scheduling.

\section{Conclusion}
\label{sec:conclusion}

In this work, we introduced PanAf-FGBG, the first wildlife dataset designed to assess backgrounds and out-of-distribution generalization in behavior recognition. Our results demonstrate that real-world background information is a strong predictor of animal behavior and can significantly enhance recognition performance. While there are some limitations — such as its restriction to statically installed camera trap videos, the additional data and compute required for latent space background subtraction, and the potential for unnatural artifacts in background concatenation — the dataset remains a valuable resource. PanAf-FGBG enables direct evaluation under out-of-distribution conditions, and its rich metadata supports additional tasks and experimental setups. We hope it will help with understanding and improving model performance in this challenging domain and ultimately benefit endangered and charismatic species like the one featured in this paper.




\section*{Acknowledgements}
\label{sec:acknowledgements}
{We thank the Pan African Programme: ‘The Cultured Chimpanzee’ team and its collaborators for allowing the use of their data for this paper. We thank Amelie Pettrich, Antonio Buzharevski, Eva Martinez Garcia, Ivana Kirchmair, Sebastian Schütte, Linda Gerlach and Fabina Haas. We also thank management and support staff across all sites; specifically Yasmin Moebius, Geoffrey Muhanguzi, Martha Robbins, Henk Eshuis, Sergio Marrocoli and John Hart. Thanks to the team at https://www.chimpandsee.org particularly Briana Harder, Anja Landsmann, Laura K. Lynn, Zuzana Macháčková, Heidi Pfund, Kristeena Sigler and Jane Widness. The work that allowed for the collection of the dataset was funded by the Max Planck Society, Max Planck Society Innovation Fund, and Heinz L. Krekeler. In this respect we would like to thank: Ministre des Eaux et Forêts, Ministère de l'Enseignement supérieur et de la Recherche scientifique in Côte d’Ivoire; Institut Congolais pour la Conservation de la Nature, Ministère de la Recherche Scientifique in Democratic Republic of Congo; Forestry Development Authority in Liberia; Direction Des Eaux Et Forêts, Chasses Et Conservation Des Sols in Senegal; Makerere University Biological Field Station, Uganda National Council for Science and Technology, Uganda Wildlife Authority, National Forestry Authority in Uganda; National Institute for Forestry Development and Protected Area Management, Ministry of Agriculture and Forests, Ministry of Fisheries and Environment in Equatorial Guinea. This work was supported by the UKRI CDT in Interactive AI (grant EP/S022937/1). This work was in part supported by the US National Science Foundation Awards No. 2118240 "HDR Institute: Imageomics: A New Frontier of Biological Information Powered by Knowledge-Guided Machine Learning" and Award No. 2330423 and Natural Sciences and Engineering Research Council of Canada under Award No. 585136 for the “AI and Biodiversity Change (ABC) Global Center”.}

\clearpage
\setcounter{page}{1}
\maketitlesupplementary

In the supplementary material, we provide: \begin{enumerate*}[label=(\roman*)] \item additional dataset statistics in Sec.~\ref{sec:add_stats}; \item a description of how synthetic background videos are created in Sec.~\ref{sec:bg_synth}; \item foreground-background video pair visualisations (see Fig~\ref{fig:full_page_fgbg}) and; \item a visualisation showcasing a small fraction ($\sim$~0.05\%) of the total available frames (see Fig.~\ref{fig:tiled_overview}) \end{enumerate*}\vspace{-6pt}

\setcounter{section}{0}
\section{Dataset Statistics}
\label{sec:add_stats}
We present additional dataset statistics for the country, research site, camera locations and behaviours. Specifically, \begin{enumerate*}[label=(\roman*)] \item the distribution of countries and the corresponding research sites are displayed in Fig.~\ref{fig:country_site_dist}; \item a comparison of the behaviour distribution for the overlapping and disjoint datasets is displayed in Fig.~\ref{fig:overlap_disjoint_diff}; \item time interval standardisation statistics are shown in Fig~\ref{fig:overlap_disjoint_diff_time} and; \item the accumulative proportion of videos contributed by each camera is shown in Fig.~\ref{fig:accum_camera_dist}.\end{enumerate*}


\setcounter{figure}{0}
\begin{figure}[!htbp]
\centering
\includegraphics[width=1.0\linewidth]{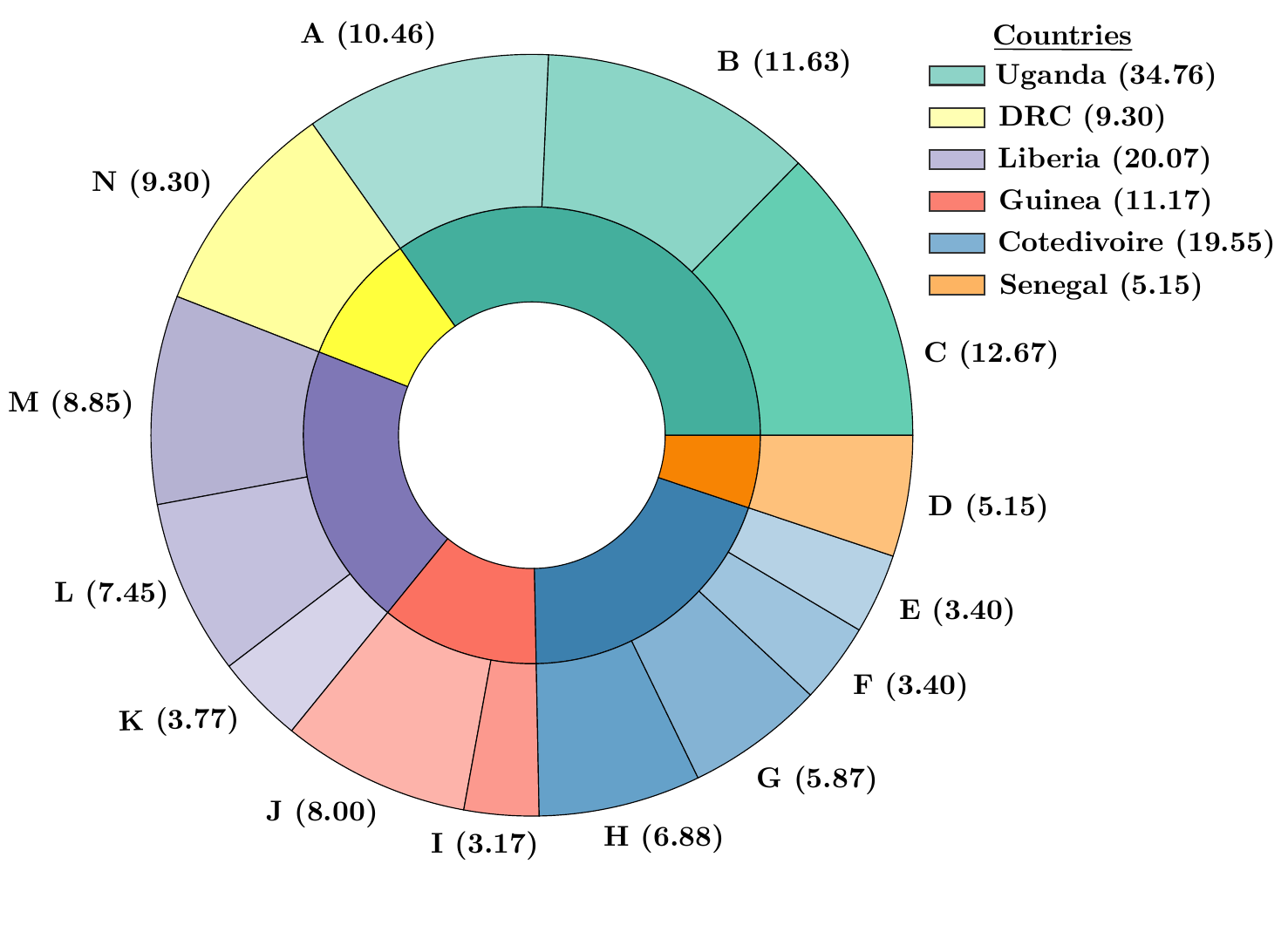}\vspace{-6pt}
\caption{{\textmd{\textbf{Proportion of videos from each country and research site}. The inner ring displays the proportion of videos extracted from each country, while the outer ring represents individual research sites. Each research site segment is a unique shade derived from its corresponding country's colour. All proportions are shown in brackets. Note that research site names are replaced with letters to protect the location of the chimps.}}}
\label{fig:country_site_dist}
\end{figure}\vspace{-6pt}

\begin{figure}[!htbp]
\centering
\includegraphics[width=1.0\linewidth, height=2.4in]{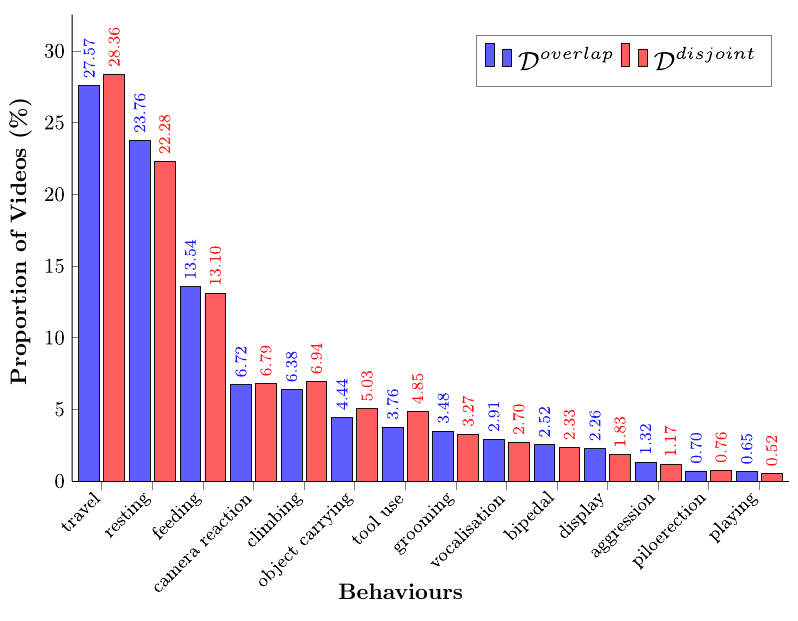}\vspace{-8pt}
\caption{{\textmd{\textbf{Comparison of the proportion of videos containing each behaviour between overlapping and disjoint datasets}. Behaviours are ordered from highest to lowest proportion, with exact values displayed above each bar for easy comparison.}}}
\label{fig:overlap_disjoint_diff}
\end{figure}\vspace{-4pt}

\begin{figure}[!htbp]
\centering
\includegraphics[width=1.0\linewidth,height=2.3in]{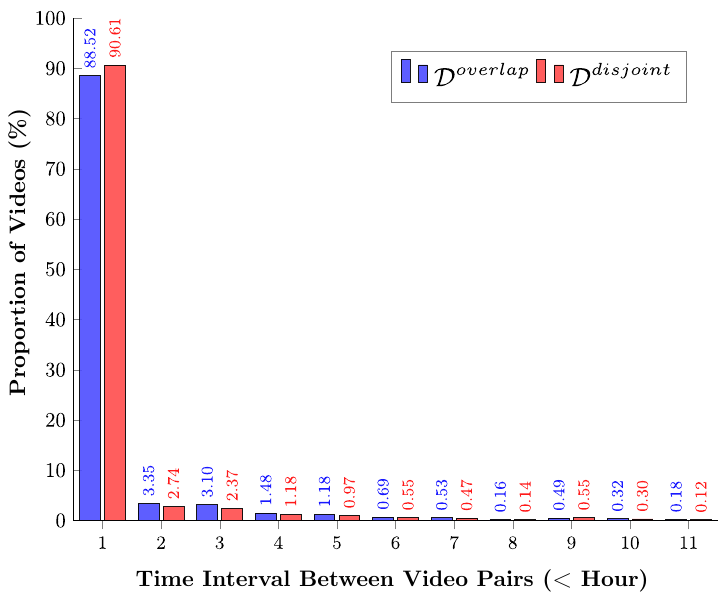}\vspace{-8pt}
\caption{{\textmd{\textbf{Time interval standardisation between foreground and background video pairs}. The majority of foreground background video pairs are sampled within one hour of each other.}}}
\label{fig:overlap_disjoint_diff_time}
\end{figure}\vspace{-4pt}


\begin{figure}[!htbp]
\centering
\includegraphics[width=1.0\linewidth, height=1.5in]{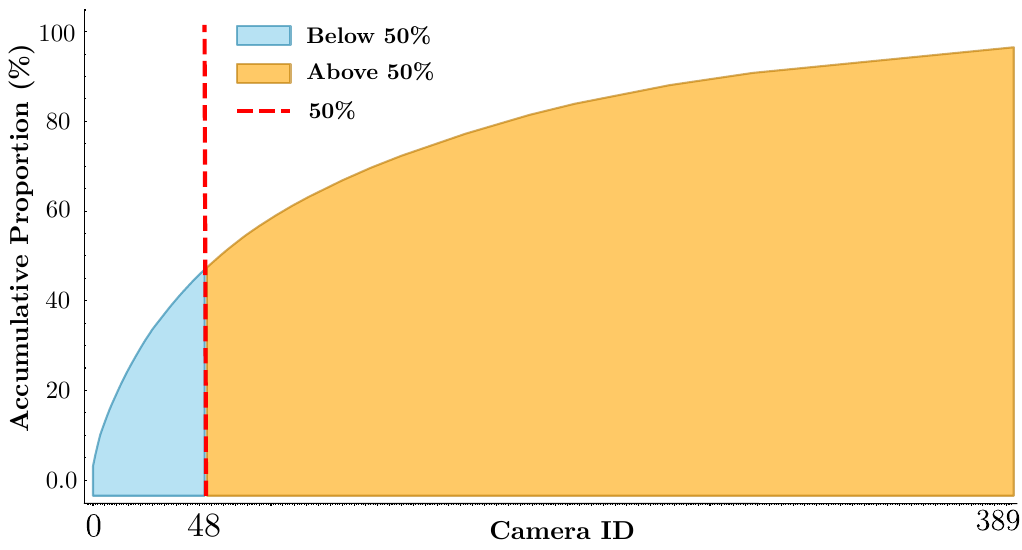}\vspace{-4pt}
\caption{{\textmd{\textbf{Accumulative proportion of videos contributed by each camera.}. The y-axis represents the accumulative proportion of videos, with individual cameras arranged by their contribution on the x-axis. The red dashed line divides the cameras into two groups: those contributing the first 50\% of the data (left) and the remaining 50\% (right).}}}
\label{fig:accum_camera_dist}
\end{figure}


\section{Synthetic Background Generation}\vspace{-2pt}
\label{sec:bg_synth}
We generated synthetic background videos using SAM2~\cite{ravi2024sam} and mean pixel value filling. Specifically, we prompted the SAM2.1-Large model using a single spatial coordinate indicating the location of the chimpanzee to produce an initial segmentation mask. We then leveraged the automatic mask propagation functionality of SAM2 to create spatio-temporal masklets for the full video. Note that spatial coordinates were produced manually. Then, we filled the area indicated by the segmentation mask with the mean pixel value for the frame (see Fig.~\ref{fig:sam2_filling} for examples).

\begin{figure*}[!ht]
\centering
\includegraphics[width=1.0\linewidth,height=0.94\textheight]{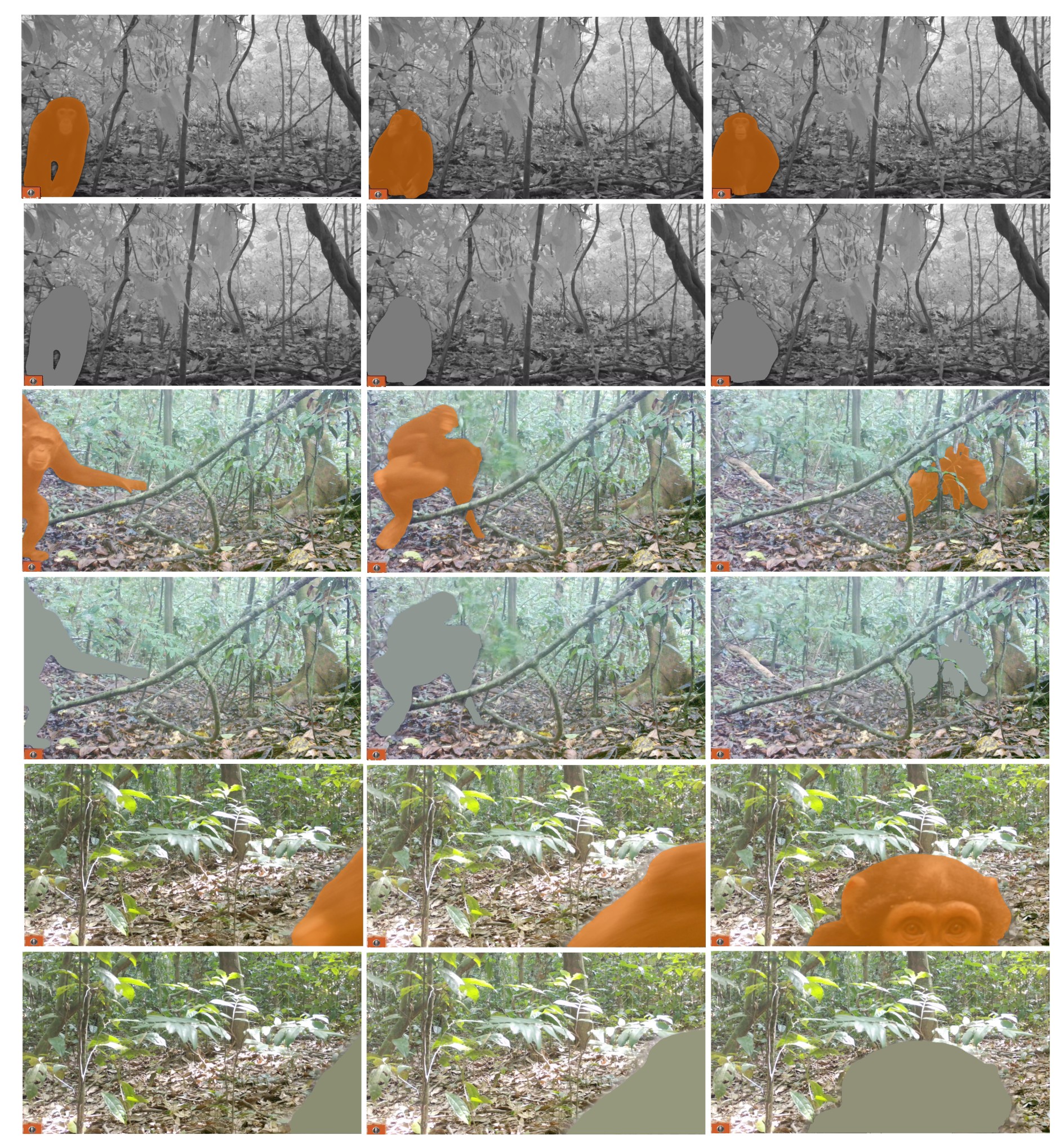}
\caption{{\textmd{\textbf{Synthetic Background Video Examples}. Three example video clips with the original segmentation mask generated by one-shot prompting of SAM2 overlaid and the corresponding mean pixel value filled frame.}}}
\label{fig:sam2_filling}
\end{figure*}

\begin{figure*}[!htbp]
\centering
\includegraphics[width=\textwidth,height=0.94\textheight]{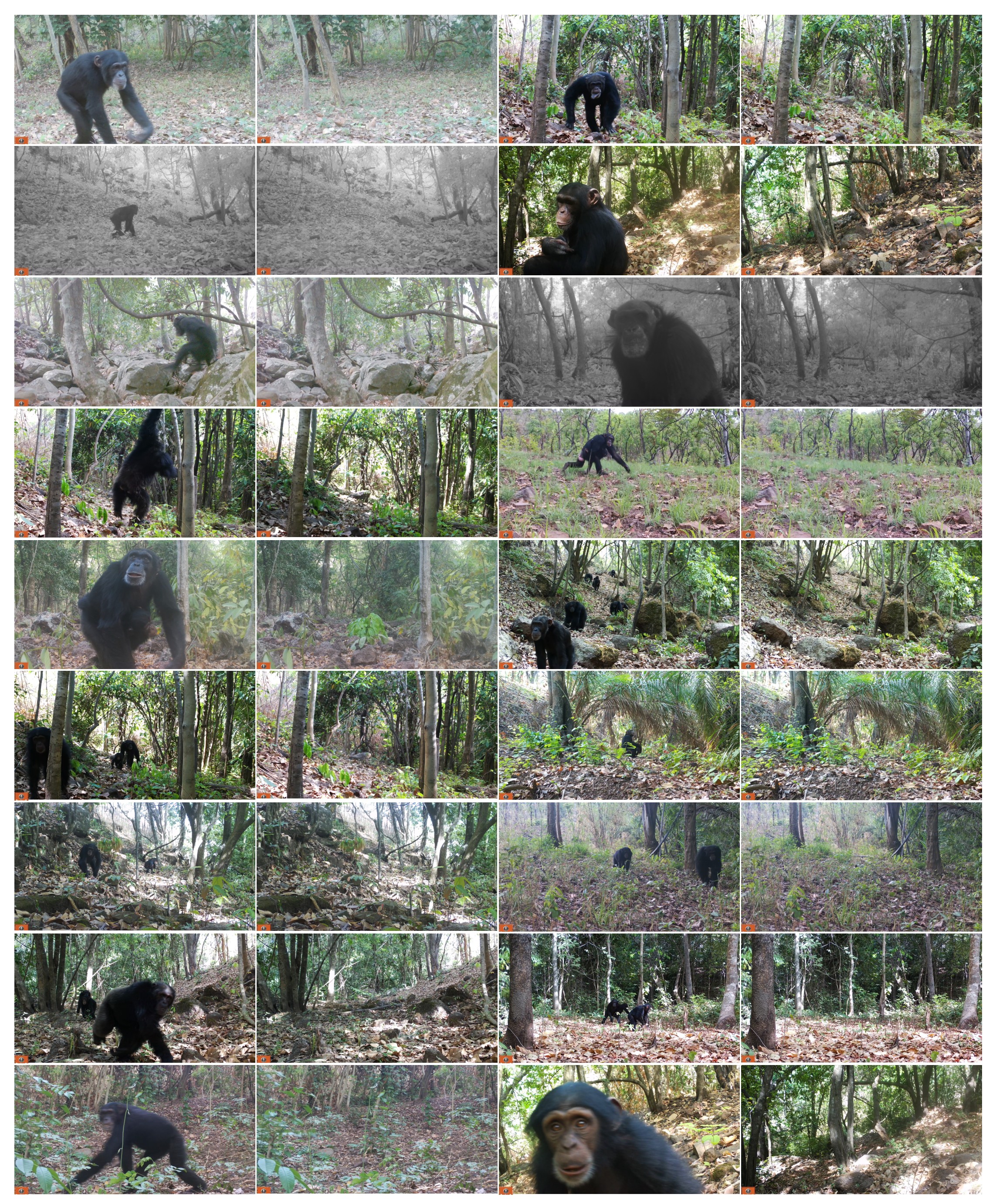}
\caption{{\textmd{\textbf{Foreground-Background Video Pair Examples}. Shown are 18 pairs of still frames (two pairs per row) extracted from foreground-background video pairs.}}}
\label{fig:full_page_fgbg}
\end{figure*}

\begin{figure*}[!htbp]
\centering
\includegraphics[width=\textwidth,height=0.94\textheight]{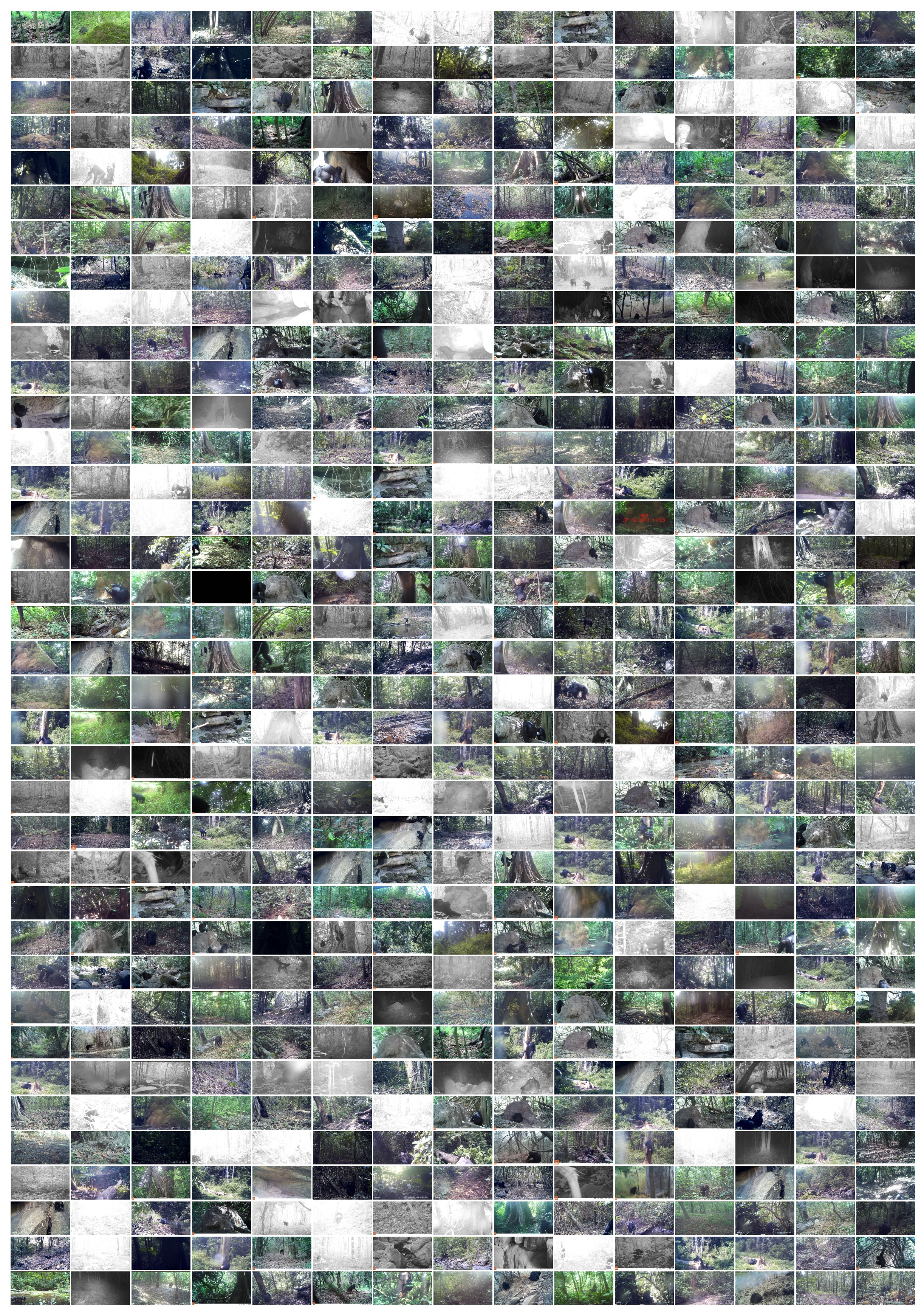}
\caption{{\textmd{\textbf{Dataset Overview}. A small fraction ($\sim$~0.05\%) of the 1.8 million frames in the dataset are shown, highlighting its diversity with respect to exhibited behaviours, habitat, weather conditions, time of day, and more.}}}
\label{fig:tiled_overview}
\end{figure*}



\clearpage

{
    \small
    \bibliographystyle{ieeenat_fullname}
    \bibliography{main}
}


\end{document}